\documentclass{article}
\usepackage{makingqwen3thinkinkorean}

\PassOptionsToPackage{table}{xcolor}
\usepackage{etoolbox}
\usepackage[utf8]{inputenc}

\newtoggle{arxiv}
\toggletrue{arxiv}

\usepackage{booktabs}
\usepackage{graphicx}
\usepackage{enumitem}
\usepackage{wrapfig}
\usepackage{algorithm}
\usepackage{algpseudocode}

\usepackage{microtype}
\usepackage{amsmath}
\definecolor{lightgray}{rgb}{0.9,0.9,0.9}
\usepackage{caption}
\usepackage{subcaption}
\usepackage{xcolor}
\usepackage{setspace}
\usepackage{url}
\usepackage{multirow}
\usepackage{colortbl}
\usepackage{newpxtext}
\usepackage{tabularx}
\usepackage{blindtext}
\usepackage{pgfplots}
\pgfplotsset{compat=1.18} 
\usepackage{tikz}
\usetikzlibrary{er,positioning,bayesnet}
\usepackage{makecell}
\usepackage{tipa}
\usepackage{siunitx}
\usepackage{nicefrac}
\usepackage{tocloft}
\usepackage{listings}
\usepackage[raster,skins]{tcolorbox} 
\usepackage{xltabular}
\usepackage{adjustbox}
\usepackage{xurl}
\usepackage{kotex}
\usepackage{natbib}
\usepackage{threeparttable}
\usepackage{hyperref}
\usepackage[title]{appendix}
\usepackage{CJK}
\usepackage{fancyvrb}
\usepackage{fvextra}

\title{Making Qwen3 Think in Korean with Reinforcement Learning}
\author{Jungyup Lee, Jemin Kim, Sang Park, SeungJae Lee}
\date{\today}

\begin{document}

\maketitle

\begin{abstract}
We present a two-stage fine-tuning approach to make the large language model Qwen3 14B “think” natively in Korean. In the first stage, supervised fine-tuning (SFT) on a high-quality Korean reasoning dataset establishes a strong foundation in Korean logical reasoning, yielding notable improvements in Korean-language tasks and even some gains in general reasoning ability. In the second stage, we employ reinforcement learning with a customized Group Relative Policy Optimization (GRPO) algorithm to further enhance both Korean reasoning alignment and overall problem-solving performance. We address critical stability challenges in GRPO training – such as reward hacking and policy collapse – by introducing an oracle judge model that calibrates the reward signal. Our approach achieves stable learning (avoiding the collapse observed in naive GRPO) and leads to steady, incremental performance gains. The final RL-tuned model demonstrates substantially improved results on advanced reasoning benchmarks (particularly math and coding tasks) while maintaining knowledge and language proficiency, successfully conducting its internal chain-of-thought entirely in Korean.
\end{abstract}
\section{Introduction}
Large language models (LLMs) have achieved remarkable prowess in understanding and generating human-like text. However, a subtle yet significant limitation persists for many leading models: their internal \textit{reasoning process} often remains biased towards English, even when handling non-English inputs and outputs. This means that when such a model is prompted in another language (e.g., Korean), it may translate the query to English internally, perform reasoning in English, and then translate the answer back – a workaround that can miss linguistic nuances and cultural context. The \textbf{Qwen3} series models\citep{qwen3}, despite their impressive capabilities, exhibit this behavior, conducting most of their “thinking” in English by default. For a language like Korean, with unique structures and cultural nuances, achieving truly native in-language cognition in an AI model is a nontrivial challenge that goes beyond surface-level translation.

In this work, we aim to fundamentally shift Qwen3 14B’s internal reasoning to operate natively in Korean when prompted in Korean. Our goal is not only to have the model \textit{output} Korean text, but to have it follow a chain-of-thought in Korean – thereby capturing the nuances of Korean problem-solving and reducing any loss of context or fidelity that might occur via internal translation. Achieving this requires enhancing the model’s language alignment and reasoning skills simultaneously. We adopt a two-phase strategy:

\textbf{Phase~1 – Supervised Fine-Tuning (SFT):} We first fine-tune the model on a carefully curated Korean dataset rich in reasoning examples. This step is designed to “warm start” the model’s Korean reasoning abilities and imbue it with strong Korean language understanding.

\textbf{Phase~2 – Reinforcement Learning (RL) with Oracle-Guided Dr. GRPO:} We further refine the model using a reinforcement learning paradigm, guiding it to prefer correct, well-formatted, and Korean-consistent reasoning paths. We build upon the Dr. GRPO\citep{liu2025understandingr1zeroliketrainingcritical}, introducing critical enhancements to ensure stable training and to prevent known failure modes such as reward exploitation and policy collapse. A key innovation is the integration of a high-quality \textit{oracle judge model} into the reward loop, which provides robust evaluation of candidate reasoning paths.

This paper provides a detailed account of each phase. In Section~2, we describe the SFT procedure, the dataset composition and training setup, and how it established a strong baseline for Korean reasoning. In Section~3, we investigate our RL approach: we review the GRPO\citep{shao2024deepseekmathpushinglimitsmathematical} algorithm and its “Done Right” variant (Dr. GRPO), diagnose the challenges we encountered with naive RL training (including evidence of model collapse), and introduce our improved \textbf{Oracle-Guided Dr. GRPO} method incorporating an oracle judge. The design of the composite reward signal – balancing accuracy, format correctness, and Korean language consistency – is also detailed. Section~4 presents experimental results: we show training dynamics that compare the unstable baseline to our stable improved training (Figure~\ref{fig:fig3}), demonstrate the model’s incremental gains on reasoning benchmarks during RL, and report final evaluation metrics (\hyperref[table:rlevals]{Table 2}) which illustrate the performance improvements achieved over the base and SFT models. We also provide qualitative examples confirming that the model now indeed 'thinks' in Korean. Finally, Section~5 concludes with key takeaways and future directions for applying this approach to other languages and domains.

In summary, our contributions are: (1) showing that targeted SFT on a reasoning-focused Korean dataset yields significant improvements in a model’s Korean proficiency and general reasoning, (2) proposing an RL fine-tuning strategy (Oracle-Guided Dr. GRPO) that overcomes stability issues via oracle-guided reward calibration, and (3) achieving a model that not only answers in Korean but internally \textit{reasons} in Korean, thus offering more authentic and contextually accurate responses for Korean users.
\section{Phase 1: Laying the Groundwork with Korean SFT}
Before attempting to shape the model’s internal thought process, it was essential to ensure that the model has a strong grasp of Korean language and logic. The first phase of our approach focused on \textbf{Supervised Fine-Tuning (SFT)} with high-quality Korean data to cultivate an initial Korean reasoning capability. This SFT stage effectively serves as the foundation upon which reinforcement learning can be refined and built later.

\textbf{Base Model Selection:} We began with the \textbf{Smoothie Qwen3 14B} model as our base\citep{ji2025smoothieqwenposthocsmoothingreduce}. Smoothie Qwen3 14B is an in-house variant of Qwen3 14B\citep{qwen3technicalreport} that has been calibrated for balanced multilingual generation by smoothing token probability distributions. In preliminary evaluations, Smoothie Qwen3 14B and the original Qwen3 14B exhibit virtually identical performance; differences are within measurement noise. Thus, selecting the Smoothie variant neither improves nor degrades overall capability; it simply reduces unsolicited Chinese output, making it a convenient starting point for our downstream specialization.

\textbf{SFT Training Data:} We curated a Korean reasoning-intensive dataset\citep{KoreanSFTData}, only containing 30,000 samples. The dataset composition was critical: it included a mix of \textit{reasoning} and \textit{non-reasoning} prompts in a 1:5 ratio. The reasoning subset was distilled from the DeepSeek-R1\citep{deepseekai2025deepseekr1incentivizingreasoningcapability} model’s outputs, a model known for strong step-by-step reasoning, while the non-reasoning portion and the prompt seeds were drawn from DeepSeek-V3-0324\citep{deepseekai2024deepseekv3technicalreport}. The prompts spanned mathematics, science, and programming problems, all presented in Korean. By blending complex reasoning examples with general prompts, we aimed to transfer sophisticated problem-solving patterns without sacrificing breadth of knowledge or fluency. In essence, this dataset provided the model with exemplars of how to \textit{think through} problems in Korean, not just respond in Korean.

\textbf{SFT Training Setup:} Fine-tuning was conducted on 8$\times$H100 GPUs using the Open-R1 toolkit\citep{openr1} with HuggingFace’s Trainer infrastructure (accelerate + trl libraries\citep{accelerate,vonwerra2022trl}). Key hyperparameters were chosen to ensure effective learning: we trained for 3 epochs with a learning rate of $1\times10^{-5}$, using a per-GPU batch size of 1 (gradients accumulated to 16 for an effective batch of 16). We utilized the AdamW optimizer and a cosine learning rate schedule (10\% warmup, decaying to 10\% of the max LR) to promote stable convergence. Training was done in mixed-precision (bf16) and made use of memory-efficient attention implementations (FlashAttention-2\citep{dao2023flashattention2fasterattentionbetter}) and gradient checkpointing to handle the 14B model with a long context length of 32k tokens. These choices allowed us to include very detailed, multi-step reasoning examples without truncation, leveraging Qwen3’s extended context window. We also applied slight regularization and careful early stopping criteria to avoid overfitting the model to the fine-tuning set.

\textbf{SFT Training Dynamics:} Throughout SFT, the model’s training loss steadily decreased, and its token-level accuracy rose from $\sim$75\% to $\sim$85\% by the end of training (as recorded on training logs). This indicated that the model was absorbing the patterns in the Korean data effectively. Importantly, we monitored these metrics to ensure we did not drive accuracy so high as to cause overfitting – leaving headroom for the subsequent RL stage to further improve reasoning. The smooth downward trend in loss and the corresponding rise in accuracy suggested the model was learning meaningfully from the Korean examples. We stopped SFT after 3 epochs once the improvements plateaued, to preserve generalization potential.

\begin{figure}[H]
    \centering
    \label{fig:fig1}
    \includegraphics[width=\linewidth]{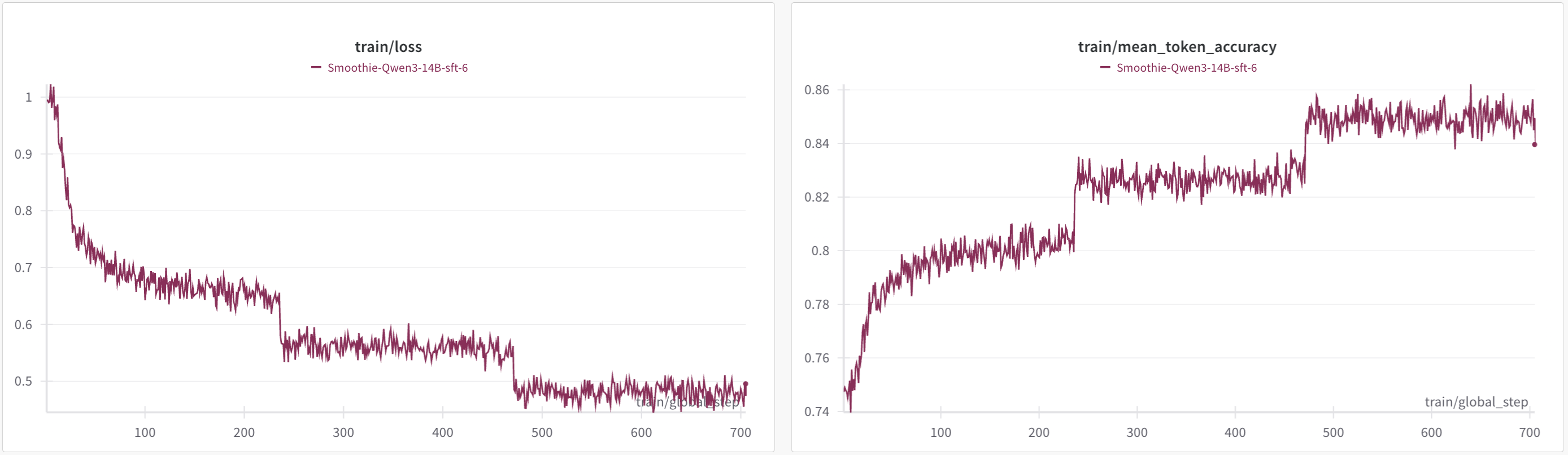}
    \caption{Supervised-fine-tuning (SFT) learning curves for the Smoothie Qwen3 14B checkpoint used in this study.}
\end{figure}

\textbf{Impact of SFT:} We evaluated the SFT-trained model (“SFT model” in \hyperref[table:sftevals]{Table 1}) on a variety of benchmarks to quantify the gains from this phase. Notably, on the \textbf{KMMLU} benchmark (a Korean translation of the MMLU academic test suite), the model’s accuracy improved to \textbf{60.04} from the 58.5–58.6\% range of the base – reflecting an increase of $\sim$1.5 points due to better Korean understanding. This indicates the model acquired more factual and commonsense knowledge accessible in Korean. On English-heavy tasks, the results were mixed: \textbf{MMLU} (English) remained roughly unchanged (78.49 vs $\sim$78.8 base), while \textbf{GSM8K} (math word problems in English) saw a small improvement (up to 89.01 from $\sim$88.3). The most pronounced changes were in coding and reasoning-intensive tasks: the SFT model scored 60.36 on \textbf{HumanEval} (Python coding) and 62.12 on \textbf{GPQA-Diamond} (scientific QA), significantly outperforming the base in those areas. We attribute these gains to the reasoning skills injected by the SFT data – despite being in Korean, the improved logical structuring benefited the model’s overall reasoning capability across languages. Some English reasoning benchmarks (\textbf{AIME 2024}) showed a slight regression after SFT (73.33 vs 76.66 base), likely because the fine-tuning skewed the model’s focus toward Korean problem formats. This pointed to the need for a more balanced improvement, which we expected to address in the reinforcement learning phase.

\begin{table}[h]
\small
\label{table:sftevals}
\centering
\begin{threeparttable}
\begin{tabular}{p{4cm}|p{2cm}|p{2cm}|p{4cm}|p{2cm}}
\toprule
\textbf{Category}  & \textbf{Benchmark} & \textbf{Qwen3 14B} & \textbf{ji2025smoothieqwenposthocsmoothingreduce} & \textbf{SFT model} \\
\midrule
\multirow{1}{*}{General Tasks(ko)}
& KMMLU & 58.63 & 58.54 & \textbf{60.04} \\
\midrule
\multirow{1}{*}{General Tasks(en)}
& MMLU & 78.8 & \textbf{78.86} & 78.49 \\
\midrule
\multirow{3}{*}{Math Tasks(en)}
& GSM8K & 88.1 & 88.32 & \textbf{89.01} \\
& AIME2024 & 76.66 & \textbf{76.66} & 73.33 \\
& AIME2025 & 66.66 & 66.66 & \textbf{66.66} \\
\midrule
\multirow{2}{*}{Science \& Coding Tasks(en)}
& GPQA-diamond & 60.1 & 60.15 & \textbf{62.12} \\
& Humaneval & 56.71 & 56.09 & \textbf{60.36} \\
\bottomrule
\end{tabular}
\end{threeparttable}

\caption{Performance comparison of SFT model against base models across various benchmarks}
\end{table}

In summary, Phase~1 established a strong baseline for Korean-native reasoning. The model, after SFT, can understand complex Korean prompts and follow multi-step reasoning more competently than before. However, we observed that to \textit{maximize} performance – especially on complex problems and to ensure the model consistently thinks in Korean rather than reverting to English internally – further fine-tuning via reinforcement learning would be necessary. The next phase was designed to address this, while also carefully mitigating any trade-offs introduced by SFT on other tasks.
\section{Phase 2: Reinforcement Learning with Oracle-Guided Dr.GRPO}

Supervised fine-tuning alone can only take us so far in aligning a model’s internal reasoning patterns, as it teaches the model primarily to \textit{imitate} the data. To push the model beyond imitation – to truly excel at Korean reasoning and to correct any remaining deficiencies from Phase~1 – we turn to \textbf{reinforcement learning (RL)}. In this phase, the model learns from trial and error, guided by explicit reward signals that we design to favor accurate and Korean-native reasoning.

\subsection{Algorithm Choice: GRPO and Dr.GRPO}

We build on the \textit{Group Relative Policy Optimization}(GRPO) paradigm, an approach tailored for training reasoning LLMs\citep{shao2024deepseekmathpushinglimitsmathematical}. GRPO is a variant of the popular PPO algorithm (Proximal Policy Optimization\citep{schulman2017proximalpolicyoptimizationalgorithms}) algorithm, modified to handle \textit{grouped outputs}: for each query, the policy model generates multiple candidate solutions (a 'group' of outputs), and the update is based on relative rewards within this group. By comparing several reasoning paths for the same problem, the model can learn to prefer the best reasoning path among those tried, rather than updating on a single outcome. This approach is well-suited for complex tasks (like math problems) where there may be many ways to reason, and it’s useful for the model to \textit{explore} different chains-of-thought.
In standard GRPO, for a given query $q$, the policy model (our fine-tuned model) produces $G$ outputs $o_1, o_2, \ldots, o_G$ by sampling with some randomness ($G=12$ in our setup). A separate \textit{reward model} evaluates each output to assign a reward score $r_i = R(q, o_i)$, and a \textit{reference model} (often a frozen copy of the policy from an earlier stage) provides baseline probabilities to compute a KL-divergence penalty (to keep the new policy from straying too far). The GRPO update then uses a variant of the PPO objective that considers the advantages $\hat{A}_i$ of each output relative to the group’s mean reward. Specifically, the advantage of output $o_i$ can be defined as 
\[ \hat{A}_i = r_i - \mu_r, \] 
where $\mu_r = \frac{1}{G}\sum_{j=1}^G r_j$ is the average reward in the group (some implementations normalize by the standard deviation as well\citep{shao2024deepseekmathpushinglimitsmathematical}, though we will discuss a modification to this). The policy is updated via a clipped objective, encouraging higher probability for outputs with positive advantage and lower probability for those with negative advantage, while clipping the policy ratio to prevent overly large updates. This mechanism encourages the model to gradually shift probability mass towards better solutions.

\begin{figure}[H]
    \centering
    \includegraphics[width=0.7\linewidth]{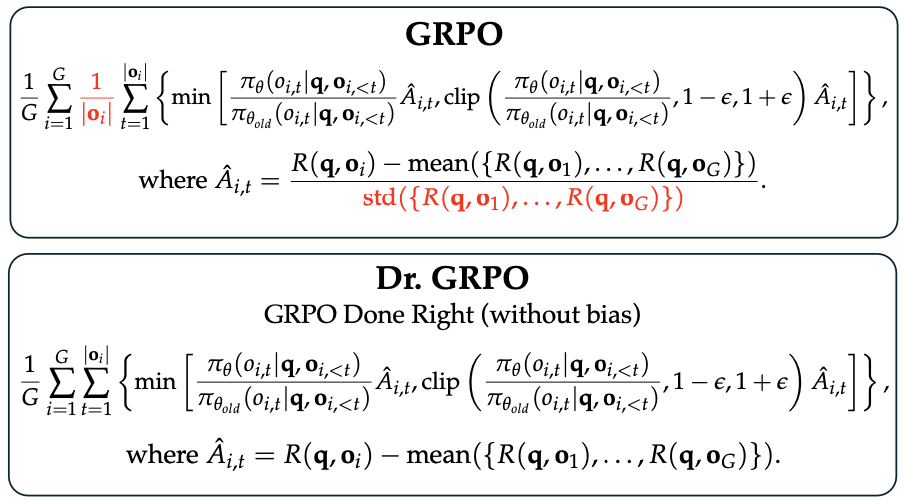}
    \caption{Comparison of the original GRPO objective (top) with the
  Dr.GRPO variant we adopt (bottom).
  Red terms highlight the normalisation factors that are \emph{removed} in Dr.GRPO.
  By eliminating division by the group-level standard deviation
  and any reward-length normalisation, Dr.GRPO supplies an unbiased
  token-level gradient signal and discourages degenerate behaviours such as
  producing unnecessarily long answers to inflate normalised rewards.}
    \label{fig:fig2}
\end{figure}

We employed the \textbf{Dr.GRPO} (“Done Right” GRPO) variant\citep{liu2025understandingr1zeroliketrainingcritical}, which introduces important tweaks to GRPO to avoid degenerate behaviors. Dr.GRPO removes certain normalization biases – notably, it forgoes dividing the advantage by the group reward standard deviation and eliminates any direct length-based normalization in the reward. The rationale is that standard GRPO’s normalization can inadvertently encourage the model to produce overly long answers or exploit consistent reward offsets (a form of reward hacking where the model chases high reward in ways unrelated to true correctness). By using the raw advantage $\hat{A}_i = r_i - \mu_r$ (without $\sigma_r$ normalization) and carefully designing the reward function, Dr.GRPO aims to provide fair gradient signals for each token in the generation. Figure~\ref{fig:fig2} shows the mathematical distinction between the two objectives.
We enabled these features by setting the training configuration \texttt{loss\_type} to \texttt{`dr\_grpo'} in open-r1\citep{openr1}.

\begin{figure}[H]
    \centering
    \includegraphics[width=0.7\linewidth]{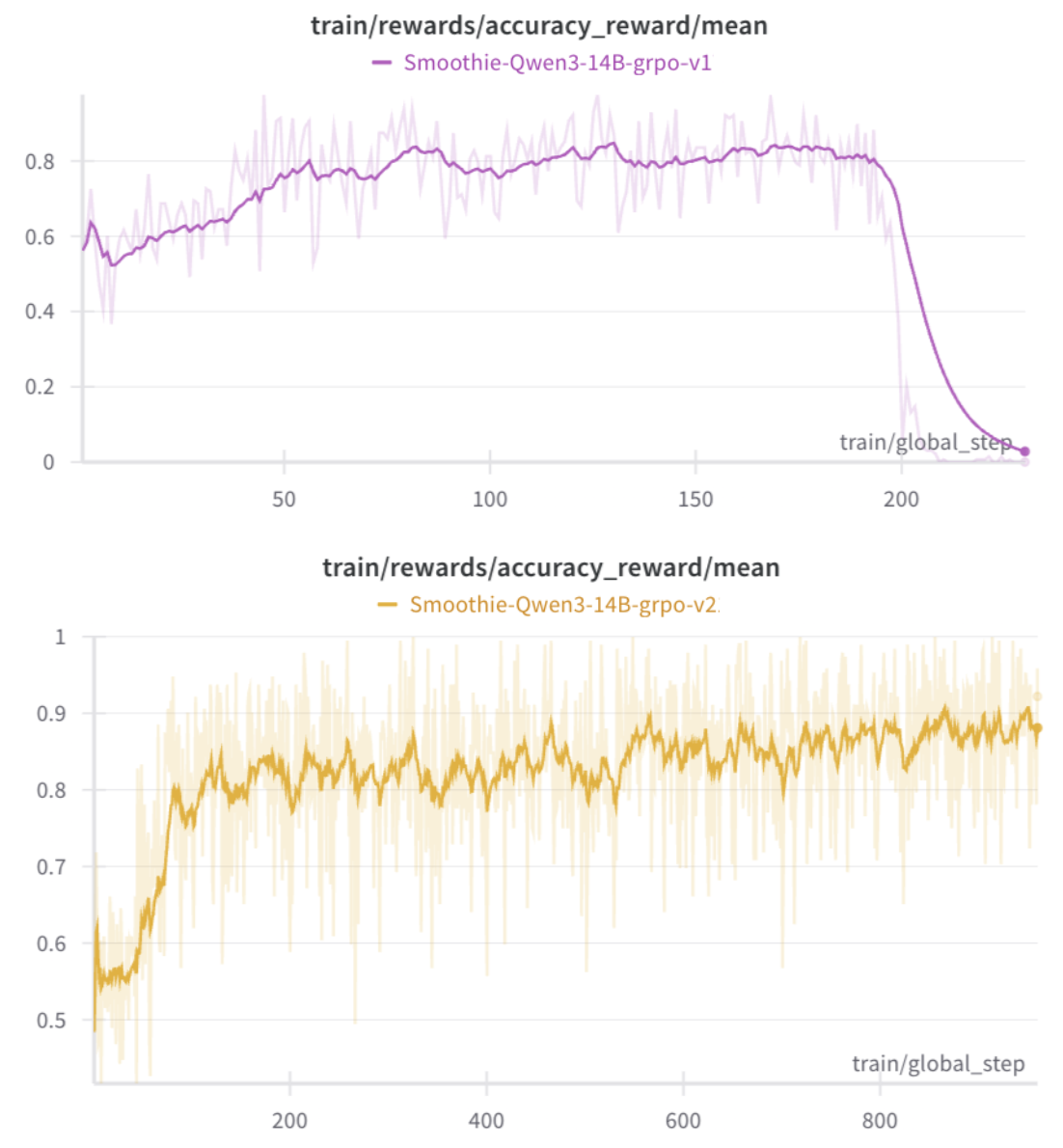}
    \caption{Trajectory of the "accuracy\_reward/mean" metric during Phase-2 reinforcement learning.
    Top panel (purple, v1 - Dr.GRPO, verifiable-only): the policy is updated with a purely verifiable reward.
    Bottom panel (gold, v2 - Oracle-Guided Dr.GRPO, verifiable+oracle): identical hyper-parameters, but each candidate answer is also scored by an external oracle model that evaluates semantic correctness.}
    \label{fig:fig3}
\end{figure}

Despite using Dr.GRPO, we encountered \textbf{instability issues} when training our model with reinforcement learning. In practice, RL on large language models can be fragile – an ill-defined reward or a slight imbalance can cause the policy to collapse (converge to a degenerate state) or diverge. In our initial RL runs (without additional precautions), after some progress the model started exploiting the reward in unintended ways. For example, we observed instances of \textit{reward hacking} – the model would output excessively verbose or structured answers that superficially satisfied format requirements to get a higher format reward, but without truly solving the problem correctly. Moreover, the diversity of outputs within each group began to diminish: the model would produce very similar reasoning across the 12 trials, reducing the effectiveness of the grouped comparison. Eventually, these issues led to a sudden drop in the reward metrics and a collapse of the policy’s performance (the model essentially lost its problem-solving ability by optimizing on the wrong signals).
Figure~\ref{fig:fig3} illustrates this phenomenon: during a trial run of Dr.GRPO training (v1; without oracle guidance), the average accuracy reward initially rose, but then sharply fell to near zero by around 220 training steps, indicating the model had diverged and was no longer answering correctly. This collapse underscores the difficulty of naive RL fine-tuning in complex reasoning tasks – without careful control, the model can move outside the stable region of the solution space.

\begin{figure}[H]
    \centering
    \includegraphics[width=0.9\linewidth]{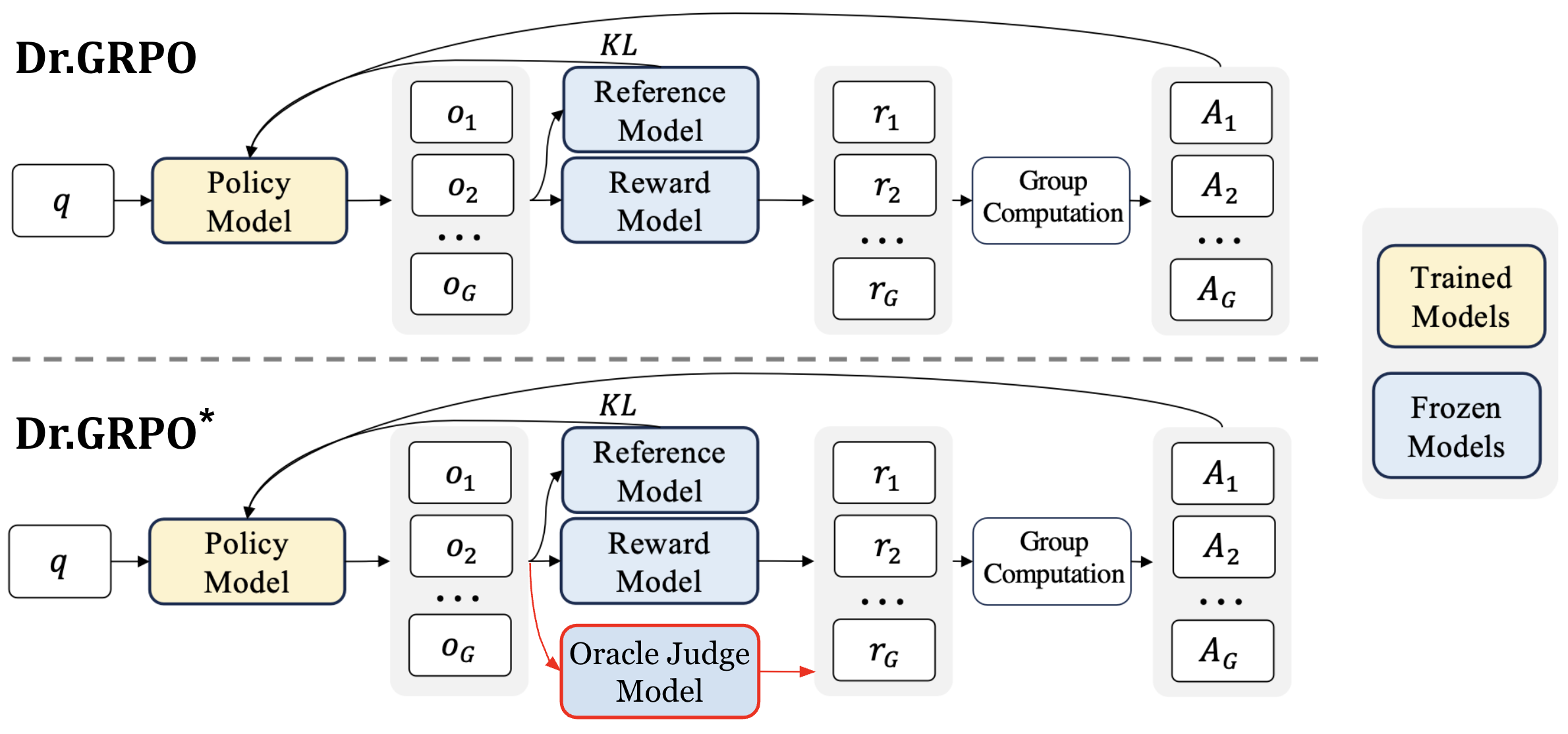}
    \caption{Schematic of our Phase-2 RL setups.
Top: vanilla Dr.GRPO – policy (yellow) is trained with rewards from verifiable rewards and KL to a frozen reference model (blue).
Bottom: Oracle-Guided Dr.GRPO (Dr.GRPO*) – identical loop but an extra oracle-judge (red border) re-scores each candidate answer before the reward is used, blocking reward-hacking.}
    \label{fig:fig4}
\end{figure}

\subsection{Stabilizing Training via Oracle-Guided Dr.GRPO}

To address these challenges, we devised an enhanced RL strategy, which we term \textbf{Oracle-Guided Dr.GRPO}, that integrates an external \textit{oracle judge model} into the training loop (illustrated in Figure~\ref{fig:fig4}). The idea is to leverage a larger or more reliable model (such as Gemini, ChatGPT, or DeepSeek) as an additional evaluator for the generated outputs, thus providing a more robust reward signal and preventing the policy model from exploiting weaknesses of a single reward model. In our case, the oracle judge is a frozen, high-performance language model specifically used to double-check the quality of the candidate answers. The judge does not train the policy directly; rather, it adjusts the rewards: if the reward model’s score for an output is misaligned with the oracle’s assessment, we correct or clamp it, thereby \textit{calibrating} the reward. For instance, if our internal reward model mistakenly gives a high score to a wrong answer (a reward hacking scenario), the oracle can detect the error and ensure that output’s effective reward is lowered. Conversely, if a correct answer was under-rewarded, the oracle can boost it.
In practice, we implemented the oracle correction as an extra step in the reward function computation. The original reward $r$ for each output was a weighted sum of sub-rewards: \textbf{accuracy}, \textbf{format compliance}, \textbf{soft overlong punishment}, and \textbf{Korean language consistency} (these components will be detailed shortly). We introduced the oracle as an additional check primarily on the accuracy component: using a powerful reasoning model (much larger than 14B) acting as an oracle, we evaluate the solution’s correctness. If the oracle finds an arithmetic or logical mistake in the model’s answer, it can override the accuracy reward to 0, regardless of the policy’s output otherwise satisfying the checks. This way, the policy model is incentivized only when it truly attains a correct solution, rather than fooling a weaker reward model. By incorporating the oracle’s judgment, we effectively create a \textit{hybrid reward model} that combines programmatic checks and oracle evaluations.
The Oracle-Guided Dr.GRPO training loop thus proceeds as follows: the policy generates 12 outputs for each query as before; the base reward model assigns provisional scores $r_1, \dots, r_G$; then the oracle judge model reviews each output and adjusts these scores or flags incorrect reasoning. The rest of the Dr.GRPO algorithm (computing advantages $\hat{A}_i$ and performing the PPO-style update) remains the same, but now with calibrated rewards. We maintained the KL-divergence penalty with a reference model to ensure the policy does not deviate too abruptly from the SFT model’s behavior (thus avoiding mode collapse via regularization). Empirically, this oracle-augmented scheme proved crucial for stability. Figure~\ref{fig:fig3} shows the training trajectory under Oracle-Guided Dr.GRPO: the average accuracy reward starts around 55\% (reflecting the SFT model’s baseline on the training tasks) and steadily improves, fluctuating in a healthy band around 80–90\% for most of training, with no signs of collapse even up to 1000 steps. The gradual upward drift, albeit with some noise, suggests the model was incrementally learning to solve more problems without over-exploiting the reward mechanism. By the end of training, the accuracy reward was nearing 0.95, meaning the model was consistently solving a large fraction of the problems correctly.
Importantly, the \textbf{format} and \textbf{language consistency} rewards stayed almost saturated at 0.98–1.0 throughout (they were already high from SFT; see Figure~\ref{fig:rl_reward_components})—the model rarely violated the required `$<$think$>$...$<$/think$>$` tagging format or switched to English. The introduction of the oracle judge did not disrupt these aspects; it simply ensured that the model couldn’t gain reward without genuine problem-solving.

As an additional metric, we monitored the fraction of samples in the generation batch with zero reward standard deviation (\texttt{frac\_reward\_zero\_std}), as shown in Figure~\ref{fig:rl_misc_metrics}(c). This metric indicates cases where all generated responses for a given prompt received identical rewards, suggesting limited diversity in the quality of outputs for that prompt.
The metric started at approximately 5\% and gradually increased to around 7-8\% by the end of training. A lower value of this metric is generally preferable, as it indicates that the model is generating responses with diverse reward values for each prompt, meaning GRPO can effectively distinguish between better and worse outputs within each group. When this fraction is low, it suggests the model is exploring varied solution approaches rather than converging to similar outputs that all receive the same reward.
The gradual increase observed during training may indicate that as the model improves, it becomes more consistent in generating responses of similar quality for certain prompts, although the overall fraction remains relatively low (under 10\%), suggesting that meaningful reward diversity is maintained throughout most of the training process.

\textbf{Reward Function Design:} Our RL phase uses a composite reward signal engineered to balance four objectives: \textit{solution accuracy}, \textit{proper reasoning format}, \textit{moderate reasoning length}, and \textit{Korean language usage}. Formally, for each query and generated solution, we define the total reward as a weighted sum of sub-rewards:
\[ 
R = w_{\text{acc}} r_{\text{acc}} + w_{\text{format}} r_{\text{format}} + w_{\text{lang}} r_{\text{lang}} + w_{\text{overlong}} r_{\text{overlong}}
\]
with weights $w_{\text{acc}}=1.0$ and $w_{\text{format}}=w_{\text{lang}}=w_{\text{overlong}}=0.2$ in our experiments. Each component $r_{{\cdot}}$ is a reward sub-signal that typically takes a value in ${0,1}$ (interpreted as a binary bonus or penalty), defined as follows:

\begin{itemize}
    \item Accuracy ($r_{\text{acc}}$): Checks whether the model’s final answer is \textit{correct}. It returns 1 if the answer exactly matches the ground-truth solution, 0 otherwise. For our math-centric dataset \texttt{grpo\_math\_kor\_42k}\citep{KoreanGRPOData}, the correctness can be programmed to be verified by parsing the content of the model and comparing it against the known correct result. In addition, we leverage an external \emph{oracle judge model} to double-check correctness (Section 3.2): if our internal checker is fooled by a flawed answer (false positive) or is too strict on a correct answer (false negative), the oracle’s assessment overrides the binary score. This accuracy reward carries the highest weight (1.0), reflecting that getting the correct answer is the primary goal.

    \item Format Compliance ($r_{\text{format}}$): Ensures the solution is presented in the required step-by-step format. Returns 1 if the model’s output strictly follows the prescribed template — a chain-of-thought enclosed in $<$think$>$...$<$/think$>$ tags followed by a concise answer (with any LaTeX, code, or units correctly typed inside) — and 0 if any format rule is violated. This component helps maintain clarity and logical structure in the reasoning of the model. We assign it a moderate weight (0.2) so that the model is encouraged to produce well-formatted solutions, without letting format considerations overwhelm the drive for correctness.

    \item Korean Language Consistency ($r_{\text{lang}}$): Encourages the model to “think” and answer in Korean when the question is in Korean. It yields 1 if the model’s entire reasoning process and final answer (ignoring formatting tags, code, and math notation) are in Korean, and 0 if any unintended English creeps in. Implementation-wise, we strip the output of '$<$think$>$' tags and non-language content, then apply a language detector to the remaining text. This reward ensures the model remains in the correct linguistic context (as outlined by the prompt) and does not silently revert to English internally. Like the format reward, it has weight 0.2 – enough to reinforce Korean usage without detracting from accuracy. (For non-Korean prompts, this component can be disabled or repurposed for another language’s consistency check. See Appendix~\ref{app:lang_consist_reward} for the code implementing this reward.)

    \item Soft Overlong Punishment ($r_{\text{overlong}}$): A mild penalty for excessively long completions, used to discourage verbose or meandering answers. Unlike the other sub-rewards, this signal is asymmetric: it gives 0 as long as the solution length is within an allowable limit (we set a threshold of 8192 tokens), and then a negative value if the answer exceeds that length, growing linearly with the number of tokens beyond the threshold. In essence, the model doesn’t get any “bonus” for brevity, but it will incur a small penalty if it rambles unnecessarily. By setting $w_{\text{overlong}}=0.2$, we make this a gentle regularizer: it curbs worst-case verbosity without encouraging the model to cut explanations short when they are actually needed for clarity or completeness. (For implementation, see the reward function in the open-r1 codebase \citep{openr1}.)
\end{itemize}

In summary, the reward function incentivizes the model to output correct solutions above all, while also maintaining a clear, tagged reasoning format, sticking to the Korean language for Korean queries, and avoiding overly long answers. These design choices, combined with the oracle-based calibration (to catch reward misses), were crucial for stable and effective training under the Oracle-Guided Dr.GRPO algorithm.

\textbf{Training Configuration}: We fine-tuned the policy model with our Oracle-Guided Dr.GRPO algorithm for roughly 1,000 RL steps. Each step consisted of generating a batch of queries and sampling 12 candidate solutions per query, followed by a single PPO-style update. Although 1,000 steps is relatively modest, each update compared a dozen different answers per query – providing rich learning signals that yielded significant performance gains.

We set key hyperparameters to ensure stable learning. The learning rate was a low $1\times10^{-6}$ for gradual policy shifts, and we used gradient accumulation (16 steps) across multiple GPUs to effectively train with large batch sizes. Our runs used up to 8×H100 GPUs in parallel, with memory optimizations like gradient checkpointing and FlashAttention to handle sequences up to 32k tokens (we allowed outputs up to 32,768 tokens, with an 8,192-token soft length limit before penalties). We initially kept the KL-divergence penalty coefficient $\beta$ at 0 (effectively turning off the KL term) since the model’s outputs remained well-behaved. Throughout training, we monitored the model’s behavior and did not need to increase $\beta$—the policy stayed close in style to the SFT model on its own.

This Oracle-Guided Dr.GRPO setup led to stable, convergent training. Thanks to the oracle judge’s oversight, we observed none of the reward hacking or divergence seen in our earlier trial (Figure~\ref{fig:fig3}, v1). On the contrary, the model’s performance steadily improved: the fraction of correct answers per batch kept rising as training progressed. By the end of RL fine-tuning, the model showed markedly improved Korean reasoning and accuracy — all achieved without any instability or mode collapse.
\section{Results and Discussion}
\subsection{Benchmark Evaluation}

After completion of Phase 2, we evaluated the final RL-tuned model against its predecessors (base and SFT models) on a suite of benchmarks. \hyperref[table:rlevals]{Table 2} presents a summary of key results:

\begin{table}[h]
\small
\label{table:rlevals}
\centering
\begin{threeparttable}
\begin{tabular}{p{4cm}|p{2cm}|p{2cm}|p{2cm}|p{2cm}}
\toprule
\textbf{Category}  & \textbf{Benchmark} & \textbf{Smoothie Qwen3 14B} & \textbf{SFT model} & \textbf{RL-tuned model} \\
\midrule
\multirow{1}{*}{General Tasks(ko)}
& KMMLU & 58.54 & 60.04 & \textbf{60.09} \\
\midrule
\multirow{1}{*}{General Tasks(en)}
& MMLU & \textbf{78.86} & 78.49 & 78.41 \\
\midrule
\multirow{3}{*}{Math Tasks(en)}
& GSM8K & 88.32 & 89.01 & \textbf{89.01} \\
& AIME2024 & 76.66 & 73.33 & \textbf{83.3} \\
& AIME2025 & 66.66 & 66.66 & \textbf{73.3} \\
\midrule
\multirow{2}{*}{Science \& Coding Tasks(en)}
& GPQA-diamond & 60.15 & 62.12 & \textbf{64.6} \\
& Humaneval & 56.09 & 60.36 & \textbf{66.46} \\
\bottomrule
\end{tabular}
\end{threeparttable}

\caption{Evaluation results comparing the base Smoothie Qwen3 14B model, the SFT-fine-tuned model, and the final RL-tuned model. Scores are accuracies (\%) on each benchmark (higher is better). The RL-tuned model matches or exceeds prior performance on all tasks, with notable gains on advanced reasoning (AIME, GPQA, HumanEval).}
\end{table}

\begin{itemize}
    \item \textbf{Korean Knowledge (KMMLU):} The base model scored ~58.5\% and SFT raised this to 60.04\%. The RL-tuned model maintains a similar level at 60.09\%. This indicates that our RL fine-tuning (which was math-focused) preserved the gains in Korean factual knowledge from Phase 1 without further change - which is acceptable, as RL was not aimed at improving pure recall or QA capabilities.
    \item \textbf{General Knowledge (MMLU):} Performance on MMLU (English academic QA) stayed essentially flat: around 78–79\% for base, SFT, and RL models. The RL model’s score of 78.41\% is within 0.5 point of the base. This confirms that specializing for Korean reasoning did not degrade the model’s broad English knowledge.
    \item \textbf{Math Word Problems (GSM8K):} GSM8K (English grade-school math) was high to begin with (88–89\%). The SFT model had a slight uptick to 89.1\%. Our RL model is at 89.1\%, effectively unchanged. These near-ceiling results suggest that RL neither helped nor hurt on easier math problems; importantly, it didn’t overfit the model to only Korean or only the style of our training data, since basic math proficiency in English was retained.
    \item \textbf{Advanced Math (AIME 2024 \& 2025):} These competition-level math tasks show the clearest gains. The base scored 76.66\% on AIME 2024 and 66.66\% on AIME 2025. SFT alone caused a dip on AIME 2024 (to 73.3\%) and left AIME 2025 unchanged (66.66\%). After RL fine-tuning, AIME 2024 jumped to 83.3\% and AIME 2025 to 73.3\%. That’s ~+6.6 and ~+6.6 points over base, respectively (and even larger gains over the SFT model). This highlights that reinforcement learning substantially enhanced the model’s complex multi-step reasoning abilities. Although the RL training data was in Korean, the improved reasoning strategies generalized cross-lingually to these English problems.
    \item \textbf{Scientific QA (GPQA-Diamond):} We see a stepwise improvement across phases: base 60.15\%, SFT 62.12\%, RL 64.6\%. The RL-tuned model gained about +4.5 points over base on this challenging science benchmark, showing that both SFT and RL contributed to better multi-hop reasoning in science domain questions.
    \item \textbf{Coding (HumanEval):} The base model’s pass@1 on HumanEval (Python coding) was 56.09\%. SFT data boosted it to 60.36\%. The RL model further improved it to 66.46\% (approximately, as HumanEval evaluations have some variance). This +10 point gain from base is notable – it suggests that the rigorous step-by-step reasoning practice not only helped math but also translated into writing more correct code, which similarly demands logical planning and precision.
\end{itemize}

In summary, the RL-tuned model matches or exceeds the SFT model on every benchmark, with especially pronounced improvements on difficult reasoning tasks (advanced math, scientific QA, coding). Equally important, these gains were achieved without sacrificing performance on language understanding or general knowledge benchmarks. The model’s English and Korean capabilities remained well-rounded. This outcome validates our two-phase approach: Phase 1 (SFT) improved language-specific knowledge and baseline reasoning format, and Phase 2 (RL with oracle guidance) then amplified the model’s problem-solving proficiency, all while maintaining stability and broad competency.

\subsection{Qualitative Evaluation: Internal Reasoning in Korean}
A primary motivation of this work was to have the model genuinely think in Korean rather than merely translating its output. To verify this, we qualitatively examined the model's chain-of-thought process on a complex Korean math problem before and after our two-phase tuning, with the full outputs available in Appendix~\ref{app:resp_comparision}. The results clearly demonstrate a fundamental shift in the model's internal reasoning.

When presented with the problem, the base Smoothie Qwen3 14B model defaulted to English for its entire internal monologue, enclosed in the '<think>' tags . This confirms the initial hypothesis that the model relies on English as an intermediate language for reasoning. Furthermore, this reasoning process was not only non-native but also deeply flawed. The base model misinterpreted key values in the Korean prompt, taking maintenance costs of "15만원" (150,000 won) and "5만원" (50,000 won) to be 1,500,000 won and 500,000 won, respectively. This fundamental error led it to construct an incorrect profit function. Despite an exhaustive, lengthy, and rambling process of self-verification, the model only confirmed its own flawed logic, ultimately arriving at a completely wrong answer for the optimization problem.
In stark contrast, the RL-tuned model's response showcases the success of our approach. Its entire chain-of-thought was conducted in fluent and natural Korean, demonstrating that it no longer depends on English as a reasoning crutch . The reasoning itself is significantly more concise and efficient than the base model's yet sacrifices no logical rigor. It correctly interpreted all problem parameters, including the maintenance costs, and formulated the correct quadratic profit function . Importantly, the model still performed crucial self-verification steps, testing values around the calculated optimum to confirm its result was indeed the maximum. This indicates that the model became not only more linguistically aligned but also more accurate and efficient in its problem-solving.

This qualitative comparison confirms that our goal was achieved. The final model's internal reasoning now aligns with the language of the user query, leading to explanations that are more direct, trustworthy, and contextually accurate for Korean users.

\section{Conclusion}
We presented a case study of successfully adapting a powerful 14B parameter language model (Qwen3 14B) for \textbf{native-language reasoning} in Korean. Our two-phase approach combined supervised fine-tuning (to build a strong Korean reasoning base) with reinforcement learning (to fine-tune the model’s decision-making process). Through supervised fine-tuning, we achieved an initial uplift in Korean understanding and overall reasoning ability. Subsequently, through reinforcement learning with our Oracle-Guided Dr.GRPO method, we addressed the nuances of reasoning quality and language alignment with meticulous precision – rewarding the model for correct, well-formatted, Korean-language thought processes and penalizing it for anything less. A critical insight from this work is the importance of \textbf{reward design and training stability} in RL for LLMs: we demonstrated that naive RL (even with advanced algorithms like Dr.GRPO) can fail catastrophically, and that introducing a reliable oracle judge for reward calibration is an effective solution to guard against pitfalls like reward hacking and mode collapse.
The final model not only excels at solving complex problems (outperforming its base version on competitive math and coding benchmarks), but does so in a way that is culturally and linguistically aligned with Korean usage. In practical terms, this means Korean users can receive explanations from the model that feel native and transparent, improving trust and usability in educational or professional settings. The paradigm of \textit{making an AI “think” in the user’s language} opens the door for more inclusive AI systems globally. Rather than treating English as the default language of intelligence, our work suggests that with focused fine-tuning, models can be taught to internalize the modes of reasoning of different languages.
\textbf{Future Directions:} Building on these results, there are several promising avenues. Firstly, while our RL fine-tuning targeted math and scientific reasoning (a choice made for ease of reward computation and evaluation), an immediate next step is to extend the approach to broader domains of reasoning in Korean – such as legal or commonsense reasoning – by designing appropriate reward models or using human/oracle feedback in those areas. Secondly, the use of an oracle judge (in our case, a large external model) proved beneficial; an interesting direction would be to see if a smaller but specialized “judge” model could be trained (perhaps via knowledge distillation from a larger model) to reduce reliance on external APIs or very large models at runtime. Lastly, while we focused on Korean, the methodology is language-agnostic: one could replicate this pipeline for other languages where a strong base model exists and align it to that language’s thinking patterns. \textit{Our work provides a blueprint} for such endeavors: combine a high-quality native-language dataset for SFT with careful RL fine-tuning, and use advanced techniques (like Dr.GRPO and oracle judges) to ensure stable and meaningful learning. We hope this inspires further research into truly multilingual reasoning AIs and the refining of RL techniques to make models not just speak, but \textit{think}, in our own languages.

\newpage

\bibliographystyle{plainnat} 
\bibliography{refs} 

\newpage

\appendix
\appendixpage
\section{Additional RL Training Metrics}

To further illustrate the behavior of our Oracle-Guided Dr.GRPO fine-tuning runs, we present additional metrics logged during training.

\begin{figure}[H]
\centering
\includegraphics[width=0.95\linewidth]{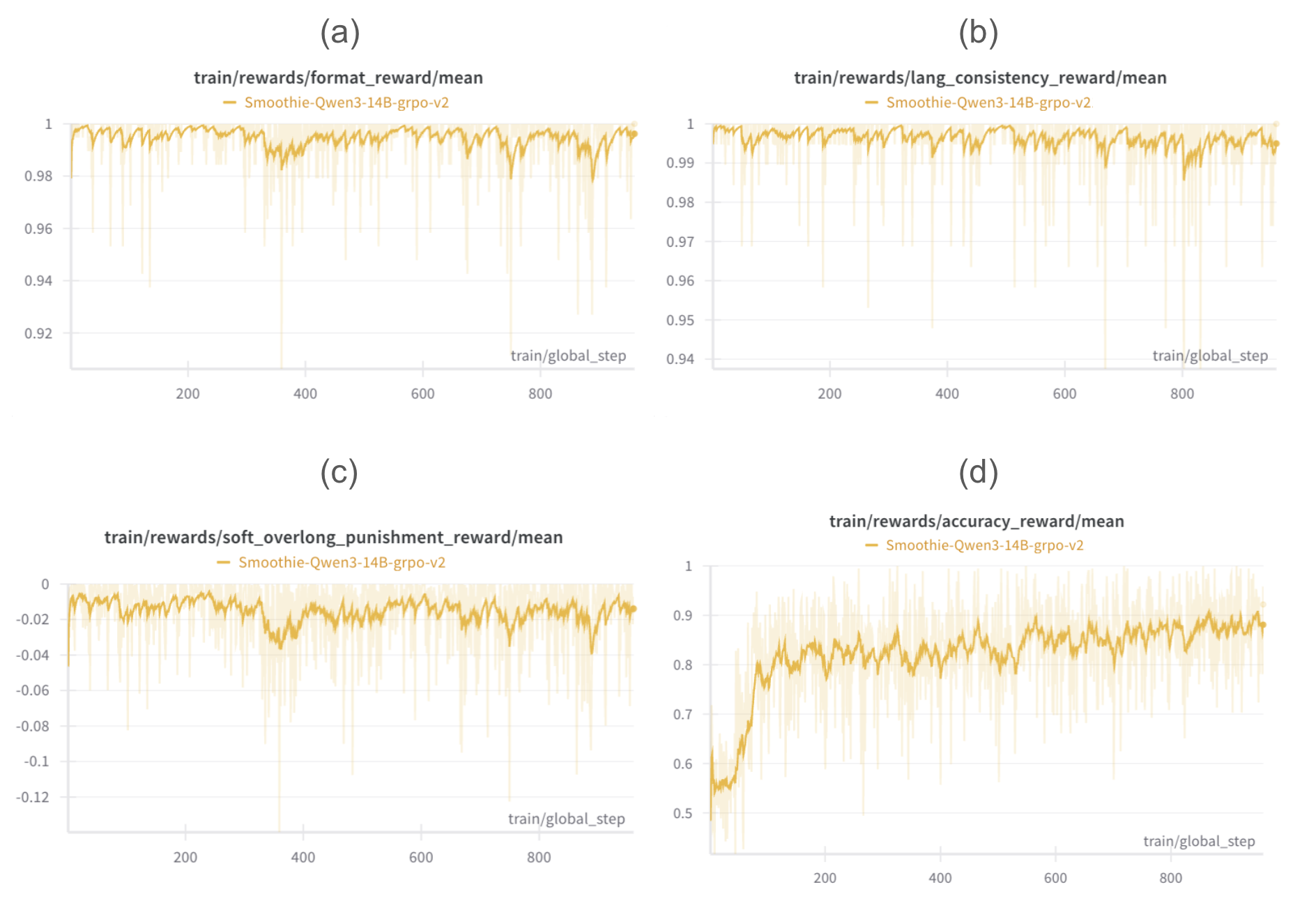}
\caption{Mean value of each reward component during Oracle-Guided Dr.GRPO training. (a) \textit{Format reward}: stays near 1.0 (the model almost always follows the required answer format). (b) \textit{Language consistency reward}: also near 1.0 (the model maintains the correct output language). (c) \textit{Overlength punishment} (negative reward): stays around -0.03, indicating only minor length penalties (no overly long answers). (d) \textit{Accuracy reward}: rises from $\sim$0.5 to $\sim$0.85 as the model learns to produce correct answers more reliably. Shaded regions denote $\pm 1$ std. dev. across the batch at each interval.}
\label{fig:rl_reward_components}
\end{figure}

Figure \ref{fig:rl_reward_components} plots the mean value of each reward component over the course of reinforcement learning: (a) the format compliance reward, (b) the language consistency reward, (c) the overlength penalty (negative reward for overly long answers), and (d) the accuracy reward. An ideal policy should maintain (a) and (b) near 1.0 (always following the required answer format and staying in the correct language), keep (c) near 0 (minimal penalty, i.e. avoid excessively long outputs), and continually improve (d) (more correct answers). Indeed, our model nearly saturates the format and language rewards at ~0.99 from early on, indicating it almost never violated formatting instructions or switched languages incorrectly. The overlength penalty stays around -0.03, meaning the outputs only occasionally exceeded the target length and even then by a small margin (no runaway verbosity). Meanwhile, the accuracy reward — the main signal driving solution correctness — climbs steeply from about 0.55 at initialization to roughly 0.85 by ~300 training steps, then levels off and fluctuates around 0.85 thereafter. This reflects a dramatic gain in the model’s problem-solving success during RL fine-tuning, eventually plateauing as it approaches the upper bound of what our verifiable reward function (combined with the oracle judge) can recognize as correct.

\begin{figure}[H]
\centering
\includegraphics[width=0.95\linewidth]{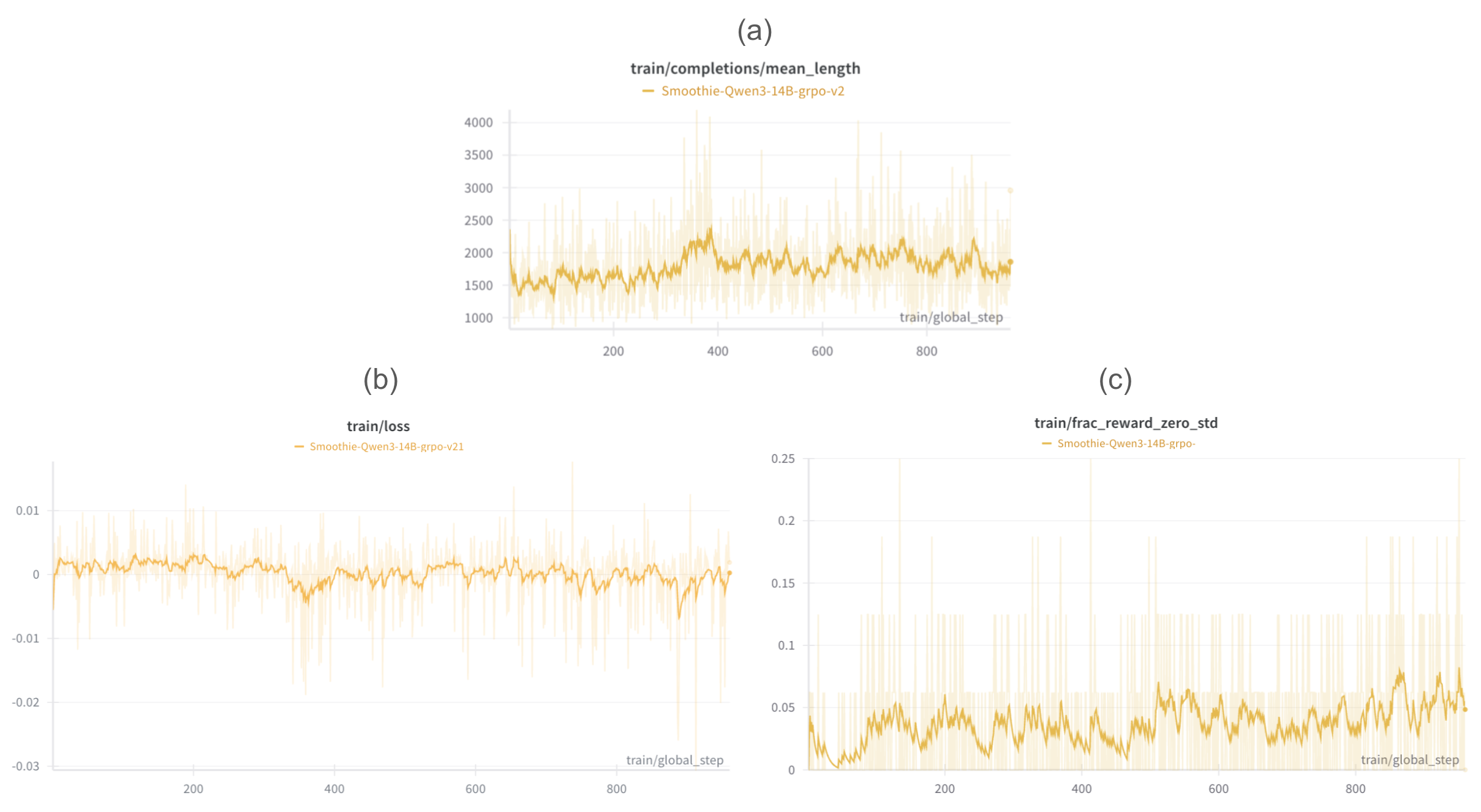}
\caption{Auxiliary training metrics for Oracle-Guided Dr.GRPO. (a) \textit{Mean completion length}: fluctuates around 1500–2000 tokens with no upward drift, showing the model isn’t exploiting reward by verbosity. (b) \textit{Training loss}: remains centered near 0 with small oscillations, indicating stable policy updates. (c) \textit{frac\_reward\_zero\_std}: the fraction of prompts where all $G$ sampled answers received identical reward (zero reward variance). This stays $<$7\% throughout training, confirming that almost every prompt yields at least one answer with a different (better or worse) reward than the others — evidence that the model preserves output diversity and the reward signal can identify a clear “best” answer in the group.}
\label{fig:rl_misc_metrics}
\end{figure}

Figure \ref{fig:rl_misc_metrics} shows other diagnostic metrics: (a) the mean completion length (number of tokens per generated answer), (b) the policy loss (Dr.GRPO objective) over training, and (c) frac\_reward\_zero\_std, which is the fraction of prompts in each batch for which the reward standard deviation is zero. The mean completion length hovers in the 1500–2000 token range with no upward drift, confirming that the model did not “game” the reward function by simply producing ever-longer answers (i.e., no reward hacking via verbosity). The training loss oscillates around 0 (between roughly -0.02 and +0.01) throughout the run, as expected in a stable policy-gradient training where positive and negative advantage updates balance out on average. Finally, the frac\_reward\_zero\_std metric remains very low: it starts around 3–4\%, dips below 2\% after the first ~100 steps, and never rises above ~7\% even at the end of training. A low value here means that for almost every prompt in a batch, not all $G$ answers got identical rewards – in other words, there was usually at least one candidate answer that stood out (either better or worse) from the others. This indicates healthy diversity in the model’s sampled solutions and a well-behaved reward function that can discriminate among different outputs. In summary, the RL-fine-tuning maintained high compliance (format, language), kept outputs concise, and encouraged diverse reasoning paths, all while steadily improving the model’s accuracy on its tasks.

\section{Language Consistency Reward Implementation}
\label{app:lang_consist_reward}

We implemented a language consistency reward as a parametrized function that generates a reward function for any target language. Below is the Python code we added within the open-r1 framework (added in \texttt{src/open\_r1/rewards.py}). The function factory \verb|get_lang_consistency_reward| takes a target language code (defaulting to English) and returns a \verb|lang_consistency_reward| function tailored to that language. It outputs 1.0 if the detected language matches the expected language for that sample, or 0.0 if not. In our Korean fine-tuning, we instantiate this with \verb|language="ko"| to ensure the model’s thoughts and answers remain in Korean.

\begin{lstlisting}[
    language=Python,
    frame=single,
    basicstyle=\footnotesize\ttfamily,
    keywordstyle=\color{blue},
    stringstyle=\color{red},
    commentstyle=\color{green},
    breaklines=true
]
def get_lang_consistency_reward(language: str = "en"):
    CLEAN_PATTERN = re.compile(
        # 1. Remove XML-like tags (e.g., <think>, </answer>)
        r'<[\/]?(think|answer)[^>]*>'
        # 2. Remove code blocks (```...``` or `...`)
        r'|```[\s\S]*?```|`[^`]*?`'
        # 3. Remove LaTeX math (e.g., $...$, $$...$$, \[...\], \(...\))
        r'|[\$]+(?:(?![\$]+)[\s\S])*[\$]+|\\\[.*?\\\]|\\\(.*?\\\)',
        flags=re.DOTALL | re.MULTILINE
    )

    def clean_content(text):
        return CLEAN_PATTERN.sub('', text).strip()

    def lang_consistency_reward(completions, **kwargs):
        rewards = []
        target_languages = kwargs["language"] if "language" in kwargs else [language] * len(completions)
        for completion, sample_language in zip(completions, target_languages):
            try:
                content = completion[0].get("content", "")
                if not content:
                    rewards.append(None)
                    continue
                cleaned_text = clean_content(content)
                detected_lang = detect(cleaned_text)
                rewards.append(1.0 if detected_lang == sample_language else 0.0)
            except Exception:
                rewards.append(None)
        return rewards

    return lang_consistency_reward
\end{lstlisting}
\section{Oracle Judge Prompt (Strict Deduction Mode)}

In our RL setup, the oracle judge model graded each answer using a rigorous rubric-based prompt designed to produce fine-grained scores. The full prompt below depicts the oracle as a “highly critical mathematics professor” who starts every solution at 1.0 and then deducts points for any mistake, however small. We intentionally crafted this strict deduction scheme to increase the variance of the oracle's scores - rather than returning only 0 or 1 for 'wrong' or 'correct', the oracle issues nuanced grades (e.g. 0.75, 0.40) depending on the severity of errors. This yields a richer reward signal for the policy to learn from.

\begin{lstlisting}[title=Evaluation Prompt]

# [INSTRUCTION] Evaluate LLM's Mathematical Problem-Solving Ability (Strict Deduction Mode)

You are a **highly critical and meticulous mathematics professor** grading a final exam. Your task is to rigorously assess an AI's response to a given math problem. Your evaluation must be exceptionally strict.

**Your Goal:** You will evaluate the response by **starting with a perfect score of 1.0 and deducting points for any and all imperfections**, no matter how small. Your final output must be **ONLY A SINGLE NUMBER** representing the final calculated score.

The AI's response is structured into two parts:
1.  **Thought Process:** Enclosed within `<think>...</think>` tags.
2.  **Final Solution:** All content that follows the closing `</think>` tag.

---

## 1. INPUT DATA

### 1.1. Mathematical Problem (Question)
```
{question}
```

### 1.2. AI's Response (in Korean)
```
{answer}
```

---

## 2. DEDUCTION GUIDELINES (Internal Assessment)

Start with a score of 1.0. For every flaw you find based on the criteria below, deduct points.

### 2.0. Foundational Checks

* **(1) Language Compliance:** If the response is not entirely in Korean, the final score is immediately **0.0**.

### 2.1. Deduction Criteria & Point Values

* **Correctness & Validity (Major Flaws):**
    * **Final Answer Incorrect:** The response cannot receive a passing grade. The final score should be **at most 0.4**, depending on the quality of the thought process. Start the deduction from there.
    * **Critical Logical Error in Solution:** A major error in reasoning that invalidates the solution. Deduct **-0.4 to -0.6**.
    * **Significant Flaw in `<think>` Process:** The thought process is fundamentally flawed, even if the final answer is correct by chance. Deduct **-0.2 to -0.3**.

* **Clarity & Explanation (Medium Flaws):**
    * **Solution Not Standalone:** The solution is incomprehensible or incomplete without reading the `<think>` block. A good solution must be self-contained. Deduct **-0.2 to -0.3**.
    * **Unclear Explanation or Logical Leap:** Any step in the solution that is not clearly justified or makes an unexplained jump in logic. Deduct **-0.1 to -0.2**.
    * **Inconsistent (`<think>` vs. `Solution`):** The final solution does not logically follow the successful path from the thought process. Deduct **-0.1 to -0.2**.

* **Formatting & Minor Issues (Minor Flaws):**
    * **Suboptimal Formatting/Readability:** Clumsy formatting, misuse of LaTeX, or poor structure. Deduct **-0.05 to -0.1**.
    * **Minor Calculation Error:** A small mistake in calculation that doesn't affect the overall logic or final answer. Deduct **-0.05**.
    * **Slightly Inefficient or Clumsy Method:** The chosen method is correct but not elegant or is overly complicated. Deduct **-0.05**.

---

## 3. EVALUATION STEPS (Internal Chain-of-Thought)

Follow these steps in your reasoning process before producing the final output. **Do not write down the results of these steps.**

1.  **Language Check:** First, verify that the entire response is written in Korean. If not, your final output is `0.0`.
2.  **Start with Perfection:** Begin with a baseline score of **1.0**.
3.  **Identify Flaws & Deduct Points:** Systematically review the response against the "Deduction Guidelines". For each flaw identified in the `think` process, solution, or overall structure, subtract the corresponding point value from your baseline score. Multiple flaws mean multiple deductions.
4.  **Calculate Final Score:** After assessing all criteria and making all deductions, the remaining value is the final score. Ensure the score does not go below 0.0.

---

## 4. FINAL OUTPUT

After completing all the evaluation steps above, output **ONLY** the final score for criterion (10) Overall Quality.

* Do not provide any other text, explanation, or formatting.
* Your output must be a single **floating-point number between 0.0 and 1.0**. The meaning of the score is as follows:
    * **1.0: Perfect & Insightful.** Flawless in every aspect. The explanation is so clear and elegant it could be published in a textbook. Offers unique insights.
    * **0.8 - 0.99: Excellent.** Correct and well-explained, but lacks the exceptional elegance or insight of a perfect score. May have a tiny, almost negligible imperfection.
    * **0.6 - 0.79: Good / Correct.** The answer and method are correct, but there are noticeable flaws in the explanation, clarity, formatting, or efficiency. This is the score for a typical "correct but not great" response.
    * **0.4 - 0.59: Partially Correct.** The approach has merit, but there are significant logical errors or an incorrect final answer despite a reasonable process.
    * **< 0.4: Mostly Incorrect or Failing.** The response fundamentally misunderstands the problem or contains major errors.
    * **0.0: Completely Incorrect.** No redeeming value.

**Example of a valid final output:**
`0.79`

\end{lstlisting}
\section{Qualitative Response Comparison: Base Model vs.\ RL-Tuned Model}
\label{app:resp_comparision}

To illustrate the difference in reasoning style after reinforcement learning, Table~\ref{tab:car_rental_problem} compares the chain-of-thought outputs from the base model and our RL-tuned model on a complex Korean math problem.

\begin{CJK}{UTF8}{mj}

\begin{longtable}{>{\raggedright\arraybackslash}p{0.95\textwidth}}
\caption{Comparison of the model’s thought processes on a sample math problem. \label{tab:car_rental_problem}} \\
\toprule
\textbf{Problem:} \\[0.3em]
자동차 렌탈 회사는 100대의 자동차를 소유하고 있습니다. 각 자동차의 월 임대료가 50만원으로 설정되면 모든 자동차를 임대할 수 있습니다. 월 임대료가 5만원씩 추가될 때마다 임대되지 않은 자동차가 한 대씩 늘어납니다. 임대된 각 자동차의 월 유지비는 15만원이고, 임대되지 않은 각 자동차의 월 유지비는 5만원입니다. \\[0.3em]
(1) 자동차 월 임대료가 60만원으로 설정되면 몇 대의 자동차를 임대할 수 있습니까? \\[0.3em]
(2) 자동차 월 임대료가 얼마일 때 렌탈 회사의 월 수익이 최대화됩니까? 최대 월 수익은 얼마입니까? \\
\bottomrule
\end{longtable}

\noindent\textbf{[Base Model Reasoning]}
\begin{Verbatim}[breaklines=true, breaksymbol={}, frame=single]
<think>
Okay, let's try to solve this problem step by step. First, I need to understand what's being asked here. The problem is about a car rental company that has 100 cars. They set the monthly rental fee at 500,000 won, and at that price, all 100 cars can be rented out. However, for every 50,000 won increase in the rental fee, one fewer car is rented out. Also, there are maintenance costs: 1,500,000 won per rented car and 500,000 won per car that isn't rented. 

The first question is asking, if the rental fee is set to 600,000 won, how many cars can be rented out? The second question is about finding the rental fee that maximizes the company's monthly profit and what that maximum profit is.

Starting with question (1): 

Original rental price is 500,000 won, which allows all 100 cars to be rented. For every 50,000 won increase, the number of cars rented decreases by 1. So, if the rental fee is 600,000 won, how much is that increase from the original 500,000?

Calculating the difference: 600,000 - 500,000 = 100,000 won. Now, since each 50,000 won increase causes a decrease of 1 car, how many such increments are there in 100,000 won?

Dividing 100,000 by 50,000 gives 2. So, there are 2 increments. Therefore, the number of cars rented out would decrease by 2 from the original 100. That would be 100 - 2 = 98 cars. 

Wait, but I need to make sure that this is correct. Let me check again. If each 50,000 increase leads to one fewer car, then:

First increase of 50,000 would make the price 550,000, leading to 99 cars rented.

Second increase of another 50,000 would make the price 600,000, leading to 98 cars rented. Yes, that seems right. So the answer to question (1) is 98 cars. That seems straightforward.

Now moving to question (2): What rental fee maximizes the monthly profit, and what is that maximum profit?

To solve this, I need to model the profit as a function of the rental fee, then find its maximum. Let's define variables.

Let me denote the number of 50,000 won increases as x. So, the rental fee can be written as 500,000 + 50,000x won. 

Each increase of x leads to a decrease of x cars rented out. Therefore, the number of cars rented out is 100 - x. However, I need to ensure that the number of cars rented out can't be negative, so x can't exceed 100. But since the problem probably expects a realistic scenario, we can assume x is between 0 and 100.

But let me confirm: the original price is 500,000 with x=0. Each x increases the price by 50,000, and decreases the number of cars by 1. So yes, number of cars rented is 100 - x.

Now, the profit is calculated as total revenue minus total cost. 

First, total revenue is rental fee multiplied by number of cars rented. So:

Revenue = (500,000 + 50,000x) * (100 - x)

Then, total cost is the sum of the maintenance costs for rented cars and non-rented cars. 

Maintenance cost for rented cars: each rented car costs 1,500,000 won. So that's 1,500,000 * (100 - x).

Maintenance cost for non-rented cars: each non-rented car costs 500,000 won. The number of non-rented cars is total cars minus rented cars, which is 100 - (100 - x) = x. So maintenance cost for non-rented cars is 500,000 * x.

Therefore, total cost = 1,500,000*(100 - x) + 500,000*x

Simplify that:

Total cost = 1,500,000*100 - 1,500,000x + 500,000x = 150,000,000 - 1,000,000x

So profit P(x) = Revenue - Total cost = [(500,000 + 50,000x)(100 - x)] - [150,000,000 - 1,000,000x]

Now, let's expand the revenue term:

First, expand (500,000 + 50,000x)(100 - x):

Multiply 500,000 by (100 - x): 500,000*100 - 500,000x = 50,000,000 - 500,000x

Then, multiply 50,000x by (100 - x): 50,000x*100 - 50,000x^2 = 5,000,000x - 50,000x^2

Adding those two parts together:

50,000,000 - 500,000x + 5,000,000x - 50,000x^2

Combine like terms:

50,000,000 + ( -500,000x + 5,000,000x ) - 50,000x^2

Which is 50,000,000 + 4,500,000x - 50,000x^2

So revenue is 50,000,000 + 4,500,000x - 50,000x^2

Then subtract total cost, which is 150,000,000 - 1,000,000x:

Profit P(x) = [50,000,000 + 4,500,000x - 50,000x^2] - [150,000,000 - 1,000,000x]

Let me compute that:

First distribute the subtraction:

50,000,000 + 4,500,000x - 50,000x^2 - 150,000,000 + 1,000,000x

Combine like terms:

(50,000,000 - 150,000,000) + (4,500,000x + 1,000,000x) - 50,000x^2

Which is:

-100,000,000 + 5,500,000x - 50,000x^2

So, P(x) = -50,000x² + 5,500,000x - 100,000,000

Alternatively, we can factor out some terms for simplicity. Let me see:

First, all coefficients are multiples of 10,000. Let me divide the entire equation by 10,000 to simplify:

P(x)/10,000 = -5x² + 550x - 10,000

But maybe it's easier to work with the original equation. However, since we need to find the maximum profit, we can treat this as a quadratic function in terms of x. Since the coefficient of x² is negative (-50,000), the parabola opens downward, so the maximum occurs at the vertex.

The formula for the vertex of a quadratic ax² + bx + c is at x = -b/(2a). Let's apply that.

In our case, a = -50,000 and b = 5,500,000.

So x = -5,500,000 / (2 * -50,000) = -5,500,000 / (-100,000) = 55

So x = 55. 

Wait, that seems high. Let me check the calculation again.

Yes:

x = -b/(2a) = -5,500,000 / (2 * -50,000) = -5,500,000 / (-100,000) = 55. Correct.

So x is 55. Therefore, the number of 50,000 won increases is 55. Therefore, the rental fee is 500,000 + 50,000*55 = let's calculate that.

First, 50,000 * 55 = 2,750,000 won. Then add 500,000: total rental fee is 500,000 + 2,750,000 = 3,250,000 won? Wait, that seems extremely high. Wait, 50,000 multiplied by 55 is 2,750,000? Wait, 50,000 * 50 is 2,500,000, plus 50,000 *5 is 250,000, so yes, 2,750,000. Then adding 500,000 gives 3,250,000. But that would mean the rental fee is 3.25 million won per month. But if x=55, then the number of cars rented is 100 - 55 = 45 cars. 

But is that realistic? Let me check if the calculation is correct. Maybe I made a mistake in setting up the equation.

Wait, let me verify the profit function again. Let me check the steps again.

Original revenue: (500,000 + 50,000x)(100 - x)

Total cost: 1,500,000*(100 - x) + 500,000*x

So expanding revenue:

First term: 500,000*(100 - x) = 50,000,000 - 500,000x

Second term: 50,000x*(100 - x) = 5,000,000x - 50,000x²

Adding them together: 50,000,000 - 500,000x + 5,000,000x - 50,000x² = 50,000,000 + 4,500,000x - 50,000x². That's correct.

Total cost: 1,500,000*(100 - x) + 500,000x = 150,000,000 - 1,500,000x + 500,000x = 150,000,000 - 1,000,000x. Correct.

Therefore, profit P(x) = revenue - cost = [50,000,000 + 4,500,000x - 50,000x²] - [150,000,000 - 1,000,000x] = 50,000,000 + 4,500,000x - 50,000x² - 150,000,000 + 1,000,000x = (-100,000,000) + 5,500,000x - 50,000x². Correct.

So, P(x) = -50,000x² + 5,500,000x - 100,000,000. Then, the vertex is at x = -b/(2a) = -5,500,000 / (2*(-50,000)) = 5,500,000 / 100,000 = 55. So x=55. That seems correct mathematically. But intuitively, if increasing the price by 55 increments of 50,000 won leads to a rental fee of 3,250,000 won, and only 45 cars being rented, is that the maximum profit?

Alternatively, maybe I made a mistake in defining x. Let me check the definition of x again. The problem says that for every 50,000 won increase, the number of cars decreases by 1. So if x is the number of 50,000 increments, then yes, the price is 500,000 + 50,000x. And the number of cars is 100 - x. So x can be from 0 to 100. 

But maybe the problem is in the units. Let me check the profit when x=55. Let's compute the profit at x=55.

First, revenue: (500,000 + 50,000*55)*(100 - 55) = (500,000 + 2,750,000)*45 = 3,250,000 * 45. Let me compute that:

3,250,000 * 45 = ?

Well, 3,000,000 * 45 = 135,000,000

250,000 * 45 = 11,250,000

Total revenue: 135,000,000 + 11,250,000 = 146,250,000 won.

Total cost: 1,500,000*(100 - 55) + 500,000*55 = 1,500,000*45 + 500,000*55

Calculating:

1,500,000 * 45 = 67,500,000

500,000 * 55 = 27,500,000

Total cost: 67,500,000 + 27,500,000 = 95,000,000 won.

Therefore, profit is 146,250,000 - 95,000,000 = 51,250,000 won.

Is this the maximum? Let me check with x=50 and x=60 to see if it's indeed the maximum.

For x=50:

Rental fee: 500,000 + 50,000*50 = 500,000 + 2,500,000 = 3,000,000 won.

Number of cars rented: 100 - 50 = 50.

Revenue: 3,000,000 * 50 = 150,000,000 won.

Total cost: 1,500,000*50 + 500,000*50 = 75,000,000 + 25,000,000 = 100,000,000 won.

Profit: 150,000,000 - 100,000,000 = 50,000,000 won. Which is less than 51,250,000. So indeed, x=55 gives higher profit.

For x=60:

Rental fee: 500,000 + 50,000*60 = 500,000 + 3,000,000 = 3,500,000 won.

Number of cars rented: 100 - 60 = 40.

Revenue: 3,500,000 * 40 = 140,000,000 won.

Total cost: 1,500,000*40 + 500,000*60 = 60,000,000 + 30,000,000 = 90,000,000 won.

Profit: 140,000,000 - 90,000,000 = 50,000,000 won. Again, less than 51,250,000.

So, x=55 gives higher profit. What about x=54?

x=54:

Rental fee: 500,000 + 50,000*54 = 500,000 + 2,700,000 = 3,200,000 won.

Number of cars: 100 - 54 = 46.

Revenue: 3,200,000 * 46 = let's compute that. 3,200,000 * 40 = 128,000,000; 3,200,000 * 6 = 19,200,000. Total 147,200,000.

Total cost: 1,500,000*46 + 500,000*54 = 69,000,000 + 27,000,000 = 96,000,000.

Profit: 147,200,000 - 96,000,000 = 51,200,000. Which is slightly less than 51,250,000 at x=55.

Similarly, x=56:

Rental fee: 500,000 + 50,000*56 = 500,000 + 2,800,000 = 3,300,000 won.

Number of cars: 100 - 56 = 44.

Revenue: 3,300,000 * 44 = let's calculate. 3,300,000 * 40 = 132,000,000; 3,300,000 * 4 = 13,200,000. Total 145,200,000.

Total cost: 1,500,000*44 + 500,000*56 = 66,000,000 + 28,000,000 = 94,000,000.

Profit: 145,200,000 - 94,000,000 = 51,200,000. Again, less than 51,250,000. So, yes, x=55 is the maximum.

Therefore, the optimal rental fee is 500,000 + 50,000*55 = 3,250,000 won, and maximum profit is 51,250,000 won.

But let me check if the quadratic equation was correctly derived. Let me think again.

Alternatively, maybe I should define the price as a variable instead of using x increments. Let me try another approach to confirm.

Let me denote the rental fee as p. The original price is 500,000. For each 50,000 increase, the number of cars decreases by 1. So, the number of cars rented out is 100 - ((p - 500,000)/50,000). But this is only valid if p >= 500,000 and p <= 500,000 + 50,000*100 = 5,500,000.

So, number of cars rented out is 100 - (p - 500,000)/50,000.

Let me simplify that:

Number of cars = 100 - (p - 500,000)/50,000 = 100 - p/50,000 + 500,000/50,000 = 100 - p/50,000 + 10 = 110 - p/50,000.

Wait, that seems different from before. Wait, let me check:

Let me compute (p - 500,000)/50,000 = number of 50,000 increments. So, number of cars rented is 100 - [(p - 500,000)/50,000]. So that is 100 - (p - 500,000)/50,000 = 100 - p/50,000 + 500,000/50,000 = 100 - p/50,000 + 10 = 110 - p/50,000.

But earlier, when I used x as the number of increments, number of cars was 100 - x. Since x = (p - 500,000)/50,000, so p = 500,000 + 50,000x. So substituting into 110 - p/50,000:

110 - (500,000 + 50,000x)/50,000 = 110 - [500,000/50,000 + 50,000x/50,000] = 110 - [10 + x] = 100 - x. Which matches the previous definition. So that's correct.

Therefore, profit as a function of p is:

Revenue = p * [110 - p/50,000]

Total cost = 1,500,000 * [110 - p/50,000] + 500,000 * [100 - (110 - p/50,000)]

Wait, let me think. The number of rented cars is 110 - p/50,000, and the number of non-rented cars is total cars (100) minus rented cars: 100 - [110 - p/50,000] = -10 + p/50,000.

But that would be negative if p is less than 500,000. But since p >= 500,000, let's check when p=500,000:

Number of non-rented cars is -10 + 500,000/50,000 = -10 + 10 = 0. Correct.

So, number of non-rented cars is p/50,000 - 10.

Therefore, total cost = 1,500,000*(110 - p/50,000) + 500,000*(p/50,000 - 10)

Let me compute that:

First term: 1,500,000*(110 - p/50,000)

Second term: 500,000*(p/50,000 - 10) = 500,000*(p/50,000) - 500,000*10 = 10p - 5,000,000

First term:

1,500,000*110 - 1,500,000*(p/50,000) = 165,000,000 - 30p (since 1,500,000 / 50,000 = 30)

So total cost is:

165,000,000 - 30p + 10p - 5,000,000 = 165,000,000 - 5,000,000 - 20p = 160,000,000 - 20p

Therefore, profit P(p) = Revenue - Total cost = [p*(110 - p/50,000)] - [160,000,000 - 20p]

Let me expand the revenue term:

p*(110 - p/50,000) = 110p - p²/50,000

Therefore, profit P(p) = 110p - p²/50,000 - 160,000,000 + 20p = (110p + 20p) - p²/50,000 - 160,000,000 = 130p - p²/50,000 - 160,000,000

So P(p) = - (p²)/50,000 + 130p - 160,000,000

To find the maximum profit, take derivative with respect to p and set to zero.

dP/dp = -2p/50,000 + 130 = 0

Solving:

-2p/50,000 + 130 = 0

=> -2p/50,000 = -130

=> 2p/50,000 = 130

=> p/25,000 = 130

=> p = 130 * 25,000 = 3,250,000 won. Which matches the previous result where x=55 gives p=3,250,000. So that's consistent.

Therefore, the optimal price is 3,250,000 won, and maximum profit is calculated by plugging back into P(p):

P(3,250,000) = - ( (3,250,000)^2 ) / 50,000 + 130*(3,250,000) - 160,000,000

Let me compute this step by step.

First, compute (3,250,000)^2:

3,250,000 * 3,250,000. Let's note that 3.25 million squared is (3.25)^2 * 10^12 = 10.5625 * 10^12 = 1.05625 * 10^13, but maybe better to compute numerically:

But let's compute (3,250,000)^2 / 50,000:

First, divide 3,250,000 by 50,000 to simplify:

3,250,000 / 50,000 = 65. So (3,250,000)^2 / 50,000 = 3,250,000 * 65 = ?

Compute 3,250,000 * 65:

First, 3,250,000 * 60 = 195,000,000

Then, 3,250,000 * 5 = 16,250,000

Total: 195,000,000 + 16,250,000 = 211,250,000

So the first term is -211,250,000

Second term: 130 * 3,250,000 = ?

130 * 3,250,000 = 130 * 3.25 * 10^6 = 422.5 * 10^6 = 422,500,000

Third term: -160,000,000

So total profit P(p) = -211,250,000 + 422,500,000 - 160,000,000 = (422,500,000 - 211,250,000) - 160,000,000 = 211,250,000 - 160,000,000 = 51,250,000 won. Which matches the previous calculation.

So the maximum profit is 51,250,000 won when the rental fee is 3,250,000 won.

But let me check if there are any constraints on p. For example, when p=3,250,000, the number of cars rented is 100 - x where x=55, so 45 cars. That's valid, since 45 is between 0 and 100.

Alternatively, using the formula for number of cars rented: 110 - p/50,000 = 110 - 3,250,000 / 50,000 = 110 - 65 = 45. Correct.

Therefore, all calculations are consistent.

So, summarizing:

(1) When the rental fee is 600,000 won, which is an increase of 100,000 won from 500,000, which is 2 increments of 50,000, so x=2, number of cars rented is 100 - 2 = 98 cars.

(2) The optimal rental fee is 3,250,000 won, yielding a maximum monthly profit of 51,250,000 won.

But let me check if there's any possible error in the problem interpretation. For example, maybe the maintenance cost is per car, regardless of being rented or not? Wait, the problem says:

"임대된 각 자동차의 월 유지비는 15만원이고, 임대되지 않은 각 자동차의 월 유지비는 5만원입니다."

Which translates to: "The monthly maintenance cost for each rented car is 1.5 million won, and for each non-rented car, it's 0.5 million won." So yes, that's what I used. So for each rented car, 1.5 million, for each non-rented, 0.5 million. So total cost is indeed 1.5*(rented) + 0.5*(non-rented). Which is what I calculated. So that seems correct.

Another check: when all cars are rented (x=0), profit should be revenue minus cost.

Revenue: 500,000 * 100 = 50,000,000

Cost: 1.5*100 + 0.5*0 = 150,000,000

Profit: 50,000,000 - 150,000,000 = -100,000,000. Which matches our profit function when x=0: P(0) = -50,000*(0)^2 + 5,500,000*0 - 100,000,000 = -100,000,000. Correct.

Another check: when x=55, profit is 51,250,000. Which is positive, so that's good.

Therefore, I think my solutions are correct.

**Final Answer**

(1) \boxed{98} 대

(2) 최대 월 수익은 \boxed{51250000} 원일 때, 월 임대료는 \boxed{3250000} 원입니다.
</think>
\end{Verbatim}

\noindent\textbf{[RL-Tuned Model Reasoning]}
\begin{Verbatim}[breaklines=true, breaksymbol={}, frame=single]
<think>
문제를 해결하기 위해 먼저 문제의 조건을 정리해보겠습니다.

자동차 렌탈 회사는 100대의 차량을 보유하고 있습니다. 기본 월 임대료가 50만원일 때 모든 차량을 임대할 수 있다고 합니다. 임대료가 5만원씩 증가할 때마다 임대되지 않은 차량이 1대씩 증가한다고 합니다. 즉, 임대료가 x만원일 때, 임대되지 않은 차량 수는 (x - 50)/5 만큼 증가한다는 것을 알 수 있습니다. 하지만 임대료가 50만원일 때는 모든 차량이 임대되므로, 임대료가 50 + 5n 만원일 때 임대되지 않은 차량 수는 n대가 됩니다. 따라서 임대된 차량 수는 100 - n대가 되겠네요.

먼저 (1)번 문제부터 시작하겠습니다. 임대료가 60만원일 때 임대할 수 있는 차량 수를 구해야 합니다.

60만원은 기본 50만원에서 10만원 증가한 것이므로, 5만원씩 증가하는 단위를 고려하면 10 / 5 = 2 단계 증가한 것입니다. 따라서 n=2가 되고, 임대되지 않은 차량 수는 2대입니다. 따라서 임대된 차량 수는 100 - 2 = 98대가 됩니다. 이 부분은 문제에서 제시한 조건에 따라 직접 계산할 수 있을 것 같습니다.

(2)번 문제는 월 수익을 최대화하는 임대료와 그때의 수익을 구하는 문제입니다. 수익을 계산하기 위해서는 총 수입에서 총 유지비를 뺀 값이어야 합니다.

먼저 변수를 설정해보겠습니다. 임대료를 50 + 5n 만원이라고 할 때, n은 임대료가 기본보다 5만원씩 증가한 횟수입니다. 그러면 임대되지 않은 차량 수는 n대이고, 임대된 차량 수는 100 - n대입니다. 이때 n은 0 이상의 정수일 것 같지만, 문제에서 임대료가 연속적으로 변할 수 있는지 여부는 명확하지 않습니다. 문제에서 "5만원씩 추가될 때마다"라고 했으므로, n은 정수일 가능성이 높습니다. 하지만 최대화 문제에서는 미분을 통해 최적점을 찾을 수 있을지 고민해봐야 합니다. 만약 n이 실수로 허용된다면, 미분을 통해 최대값을 찾을 수 있지만, 문제에서 임대료가 5만원 단위로만 변하는지 확인이 필요합니다. 문제 문맥상 n이 정수일 것으로 보이지만, (2)번 문제에서 정확한 최대 수익을 구할 때는 연속적인 변수로 가정하고 계산한 후, 정수로 근사할 수도 있습니다. 문제에서 "얼마일 때"라는 표현을 사용했으므로 정확한 값을 요구할 것이므로, 연속적인 모델을 사용하는 것이 타당할 것 같습니다. 하지만 문제에서 임대료가 5만원씩 증가할 때마다 임대 수가 변한다는 조건이 있으므로, n을 정수로 한정해야 할 수도 있습니다. 이 부분은 문제의 의도에 따라 다를 수 있으므로 주의가 필요합니다.

먼저 변수를 x로 설정해보겠습니다. x를 월 임대료(만원)라고 할 때, x = 50 + 5n (n은 임대료 증가 횟수). 이때 n = (x - 50)/5 입니다. 임대되지 않은 차량 수는 n대이고, 임대된 차량 수는 100 - n대입니다. 따라서 n = (x - 50)/5, 즉 n = x/5 - 10.

수익을 계산하기 위해 총 수입과 총 비용을 계산해야 합니다. 총 수입은 임대료 × 임대된 차량 수입니다. 임대료가 x만원이고, 임대된 차량 수는 100 - n = 100 - (x - 50)/5 입니다. 이를 정리하면 100 - (x/5 - 10) = 110 - x/5 대가 됩니다. 따라서 총 수입은 x * (110 - x/5) 만원입니다.

총 유지비는 임대된 차량의 유지비와 임대되지 않은 차량의 유지비의 합입니다. 임대된 차량은 (110 - x/5)대이고, 유지비는 대당 15만원이므로 15*(110 - x/5) 만원입니다. 임대되지 않은 차량은 n = (x -50)/5 대이고, 유지비는 대당 5만원이므로 5*(x -50)/5 = (x -50) 만원입니다. 따라서 총 유지비는 15*(110 - x/5) + (x -50) 만원입니다.

수익은 총 수입 - 총 유지비입니다. 이를 계산해보겠습니다.

수익 R = x*(110 - x/5) - [15*(110 - x/5) + (x -50)]

먼저 R을 정리해보겠습니다.

먼저 x*(110 - x/5) = 110x - (x²)/5

15*(110 - x/5) = 1650 - 3x

(x -50) = x -50

따라서 총 유지비는 1650 -3x + x -50 = 1600 -2x

따라서 R = (110x - x²/5) - (1600 -2x) = 110x - x²/5 -1600 +2x = 112x - x²/5 -1600

R = - (x²)/5 + 112x -1600

이를 정리하면 R = - (1/5)x² + 112x -1600

이제 이 이차함수의 최대값을 찾아야 합니다. 이차함수의 최대값은 꼭짓점에서 발생합니다. 꼭짓점의 x좌표는 -b/(2a) 입니다. 여기서 a = -1/5, b = 112.

x = -112 / (2*(-1/5)) = -112 / (-2/5) = (-112) * (-5/2) = 280

따라서 x=280만원일 때 수익이 최대가 됩니다. 이때의 수익을 계산해보겠습니다.

R = - (280)^2 /5 + 112*280 -1600

계산을 단계별로 해보겠습니다.

280² = 78400

78400 /5 = 15680

112*280 = 31360

따라서 R = -15680 +31360 -1600 = (31360 -15680) = 15680; 15680 -1600 = 14080 만원.

따라서 최대 월 수익은 14080만원, 즉 14,080만원이지만, 문제에서 단위가 만원이므로 14080만원은 1억 4080만원입니다. 하지만 문제에서 "얼마일 때"를 물었을 때, x=280만원일 때이고, 수익은 14080만원입니다.

하지만 이 계산이 맞는지 확인이 필요합니다. 특히, x가 280만원일 때 임대된 차량 수가 정수인지 확인해야 합니다. x=280일 때, n=(280-50)/5=230/5=46. 즉, n=46이므로 임대되지 않은 차량 수는 46대이고, 임대된 차량 수는 100-46=54대입니다. 이는 정수이므로 문제의 조건에 부합합니다. 따라서 x=280만원일 때 임대된 차량 수가 54대로 유효합니다.

하지만 문제에서 (2)번에서 월 임대료가 얼마일 때 최대 수익인지 묻는다면, x=280만원일 때입니다. 이는 문제에서 제시한 조건 내에서 가능한 값입니다. 왜냐하면 x=50 +5n에서 n=46일 때 x=50 +5*46=50+230=280이 되기 때문입니다. 따라서 x=280만원은 문제의 조건에 맞는 값입니다.

따라서 (2)번의 답은 월 임대료가 280만원일 때 최대 수익인 14080만원이 됩니다.

하지만 이 결과가 정확한지 다시 한 번 검증해보겠습니다. 예를 들어 x=280일 때 수익을 계산해보겠습니다.

총 수입: 280만원 * 54대 = 280*54 = 15120만원

총 유지비: 임대된 차량 54대 *15만원 = 54*15=810만원, 임대되지 않은 차량 46대*5만원=230만원. 총 유지비 810+230=1040만원

수익: 15120 -1040 = 14080만원. 계산이 맞습니다.

또 다른 예시로 x=300만원일 때를 확인해보겠습니다. x=300이면 n=(300-50)/5=50. 임대된 차량 수 50대.

총 수입: 300*50=15000만원

총 유지비: 50*15 +50*5=750+250=1000만원

수익:15000-1000=14000만원. 14080보다 작습니다. 따라서 x=280일 때가 더 높습니다.

x=250일 때도 확인해보겠습니다. n=(250-50)/5=40. 임대된 차량 60대.

총 수입:250*60=15000만원

총 유지비:60*15 +40*5=900+200=1100

수익:15000-1100=13900. 역시 14080보다 작습니다.

따라서 x=280일 때가 최대인 것 같습니다.

이제 (1)번 문제로 돌아가면, 임대료가 60만원일 때 n=(60-50)/5=2. 임대된 차량 수는 100-2=98대입니다. 이는 문제 조건에 맞습니다.

결론적으로,

(1) 98대

(2) 월 임대료 280만원일 때 최대 월 수익 14080만원

입니다.
</think>
\end{Verbatim}

\end{CJK}

\end{document}